\documentclass{sig-alternate-05-2015}
\usepackage{graphicx} 
\usepackage{epstopdf}

\usepackage{multirow}
\usepackage{amsmath,amssymb}
\usepackage[noend,lined,boxed,vlined,ruled,linesnumbered]{algorithm2e}
\usepackage{caption}
\usepackage{tikz}
\usepackage{pgflibraryarrows}
\usepackage{pgflibrarysnakes}
\usepackage{subfigure}
\usepackage[utf8]{inputenc}
\usepackage[english]{babel}
\usepackage{times}
\usepackage{cite}

\newtheorem{definition}{Definition}

\newtheorem{property}{Property}

\newcommand{\ttfont}[1]{\ensuremath{{\tt #1}}\xspace}

\newcommand{\name}  {HINE}

\newcommand{\eat}[1]{{}}

\makeatletter
  \newcommand\figcaption{\def\@captype{figure}\caption}
  \newcommand\tabcaption{\def\@captype{table}\caption}
\makeatother

\begin{document}

\title{Heterogeneous Information Network Embedding for \\Meta Path based Proximity}

\numberofauthors{1}
\author{Zhipeng Huang, Nikos Mamoulis
\vspace{.2em} \\ \affaddr{The University of Hong Kong}
\vspace{.2em} \\ \email{\{zphuang,~nikos\}@cs.hku.hk}
}

\maketitle
\thispagestyle{plain}
\pagestyle{plain}

\begin{abstract}
A network embedding is a representation of a large graph in a low-dimensional space, 
where vertices are modeled as vectors. 
The objective of a good embedding is to preserve the proximity (i.e., similarity) 
between vertices in the original graph.
This way, typical search and mining methods 
(e.g., similarity search, kNN retrieval, classification, clustering) can be applied in the embedded space
with the help of off-the-shelf multidimensional indexing approaches. 
Existing network embedding techniques focus on homogeneous networks, where all vertices are considered to belong to a single class.
Therefore, they are weak in supporting similarity measures for heterogeneous networks.
In this paper, we present an effective heterogeneous network embedding
approach for meta path based proximity measures.   
We define an objective function, which aims at minimizing the distance between two distributions, one modeling the meta path based proximities, the other modeling the proximities in the embedded vector space. 
We also investigate the use of negative sampling to accelerate the optimization process. As shown in our extensive experimental evaluation, our method creates embeddings of high quality and  
has superior performance in several data mining tasks
compared to state-of-the-art network embedding methods.
\end{abstract}

\keywords{heterogeneous information network; meta path; network embedding}

\section{Introduction}

The availability and growth of large networks, such as social networks, co-author networks, and knowledge base graphs, has given rise to numerous applications that search and analyze information  in them.
However, for very large graphs, common information retrieval and mining tasks such as link prediction, node classification, clustering, and recommendation are time-consuming.
This motivated a lot of interest \cite{grarep, deepwalk, line} in approaches that {\em embed} the network into a low-dimensional space, such that the original vertices of the graph are represented as vectors. A good embedding preserves the proximity (i.e., similarity)  between vertices in the original graph. Search and analysis can then be applied on the embedding with the help of efficient algorithms and indexing approaches for vector spaces.

\begin{figure}
 \begin{minipage}{2\linewidth}
 \includegraphics[scale=0.75]{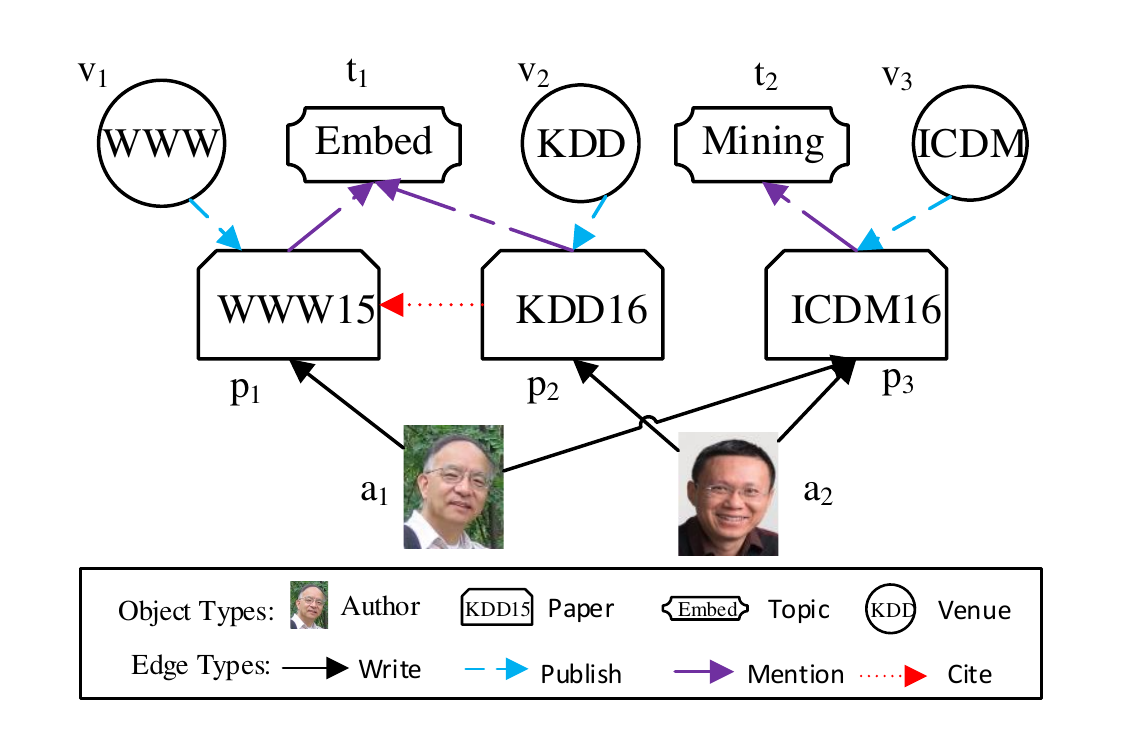}
\end{minipage}
  \caption{An example of HIN}
  \label{fig:hinExp}
\end{figure}

Heterogeneous information networks (HINs), such as DBLP \cite{ley2005dblp}, YAGO \cite{suchanek2007yago}, DBpedia \cite{auer2007dbpedia} and Freebase \cite{bollacker2008freebase}, are networks with nodes and edges that may belong to multiple types. These graph data sources contain a vast number of interrelated facts, and they can facilitate the discovery of interesting knowledge~\cite{jayaram2013querying,pcrw,fspg,mottin2014exemplar}. Figure~\ref{fig:hinExp} illustrates a HIN, which describes the relationships between objects (graph vertices) of different types (e.g., author, paper, venue and topic). For example, \ttfont{Jiawei~Han} ($a_1$) has {\it written} a \ttfont{WWW} paper ($p_{1}$), which {\it mentions} topic ``\ttfont{Embed}'' ($t_1$).

Compared to homogeneous networks, the relationships between objects in a HIN are much more complex. The proximity among objects in a HIN is not just a measure of closeness or distance, but it is also based on {\it semantics}. For example, in the HIN in Figure \ref{fig:hinExp}, author $a_1$ is close to both $a_2$ and $v_1$, but these relationships have different semantics. $a_2$ is a co-author of $a_1$, while $v_1$ is a venue where $a_1$ has a paper published. 

{\it Meta path} \cite{pathsim} is a recently proposed proximity model in HINs. A meta path is a sequence of object types with edge types in between modeling a particular relationship. For example, $A \rightarrow P \rightarrow V \rightarrow P \rightarrow A$ is a meta path, which states that two authors ($A$) are related by their publications in the same venue ($V$). Based on meta paths, several proximity measures have been proposed.
For example, \emph{PathCount} \cite{pathsim} counts the number of meta path instances connecting the two objects, while \emph{Path Constrained Random Walk (PCRW)}~\cite{pcrw} measures the probability that a random walk starting from one object would reach  the other via a meta path instance.
These measures have been shown to have better performance compared to proximity measures not based on meta paths, in various important tasks, such as $k$-NN search \cite{pathsim}, link prediction \cite{sun2011co,sun2012will,zhang2014meta}, recommendation \cite{yu2014personalized}, classification \cite{ji2011ranking,kong2012meta} and clustering \cite{sun2013pathselclus}.

Although there are a few works on embedding HINs \cite{chang2015heterogeneous,tri}, 
none of them 
is designed for meta path based proximity in general HINs. 
To fill this gap, in this paper, we propose \name, which learns a transformation of the objects (i.e., vertices) in a HIN to a low-dimensional space, such that the meta path based proximities between objects are preserved. More specifically, we define an appropriate objective function that preserves the relative proximity based rankings the vertices in the original and the embedded space.
As shown in \cite{pathsim}, meta paths with too large lengths are not very informative; therefore, we only consider meta paths up to a given length threshold $l$. 
We also investigate the use of negative sampling \cite{mikolov2013distributed} in order to accelerate the optimization process.



We conduct extensive experiments on four real HIN datasets to compare our proposed \name\ method with state-of-the-art network embedding methods (i.e., LINE \cite{line} and DeepWalk \cite{deepwalk}), which do not consider meta path based proximity.
 Our experimental results show that our \name\ method with PCRW as the meta path based proximity measure outperforms all alternative approaches in most of the qualitative measures used.


The contributions of our paper are summarized as follows:
\begin{itemize}

\item This is the first work that studies the embedding of HINs for the preservation of meta path based proximities. This is an important subject, because meta path based proximity has been proved to be more effective than traditional structured based proximities.

\item We define an objective function, which explicitly aims at minimizing the the distance between two probability distributions, one modeling the meta path based proximities between the objects, the other modeling the proximities in the low-dimensional space.

\item We investigate the use of negative sampling to accelerate the process of model optimization.

\item We conduct extensive experiments on four real HINs, which demonstrate that our \name\ method is the most effective approach for preserving the information of the original networks; \name\ outperforms the state-of-the-art embedding methods in many data mining tasks, e.g., classification, clustering and visualization.

\end{itemize}

The rest of this paper is organized as follows. Section \ref{sec:related} discusses related work. In Section \ref{problemdefinition}, we formally define the problem. Section \ref{method} describes our \name\ method. In Section \ref{section:experiments}, we report our experimental results. Section \ref{conclusion} concludes the paper.

\section{Related Work}\label{sec:related}

\subsection{Heterogeneous Information Networks}\label{sec:related:hin}
The heterogeneity of nodes and edges in HINs bring challenges, but also opportunities 
to support important applications.
Lately, there has been an increasing interest in both academia and industry in 
the effective search and analysis of information from HINs.  
The problem of classifying objects in a HIN by authority propagation is studied in \cite{ji2011ranking}.
Follow-up work \cite{kong2012meta} investigates 
a collective classification problem in HINs using meta path based dependencies. 
PathSelClus \cite{sun2013pathselclus} is a link based clustering algorithm for HINs, 
in which a user can specify her clustering preference by providing some examples as seeds.
The problem of link prediction on HINs has been extensively studied \cite{sun2011co,sun2012will,zhang2014meta}, due to its important applications (e.g., in recommender systems). 
A related problem is entity recommendation in HINs \cite{yu2014personalized}, 
which takes advantage of the different types of relationships in HINs to provide better recommendations.

\subsection{Meta Path and Proximity Measures}\label{sec:related:path}

Meta path \cite{pathsim} is a general model for the proximity between objects in a HIN. 
Several measures have been proposed for the proximity between objects w.r.t. a given meta path $\mathcal{P}$. PathCount measures the number of meta path instances connecting the two objects, and PathSim is a normalized version of it \cite{pathsim}. Path constrained random walk (PCRW) was firstly proposed \cite{pcrw} for the task of relationship retrieval over bibliographic networks. Later, \cite{fspg} proposed an automatic approach to learn the best combination of meta paths and their corresponding weights based on PCRW. Finally, HeteSim \cite{hetesim} is recently proposed as an extension of meta path based SimRank. In this paper, we focus on the two most popular proximity measures, i.e., PathCount and PCRW.

\subsection{Network Embedding}\label{sec:related:embed}

Network embedding aims at learning low-dimensional representations for the vertices of a network, such that the proximities among them in the original space are preserved in the low-dimensional space.
Traditional dimensionality reduction techniques \cite{cox2000multidimensional, tenenbaum2000global, roweis2000nonlinear, belkin2001laplacian} typically construct the affinity graph using the feature vectors of the vertexes and then compute the eigenvectors of the affinity graph. 
Graph factorization \cite{ahmed2013distributed} finds a low-dimensional representation of a graph through matrix factorization, after representing the graph as an adjacency matrix. However, since these general techniques are not designed for networks, they do not necessarily preserve the global network structure, as pointed out in \cite{line}.


Recently, DeepWalk \cite{deepwalk} is proposed as a method for learning the latent representations of the nodes of a social network, from truncated random walks in the network. 
DeepWalk combines random walk proximity with the SkipGram model \cite{skipgram}, a language model that maximizes the co-occurrence probability among the words that appear within a window in a sentence. However, DeepWalk has certain weaknesses when applied to our problem settings. First, the random walk proximity it adopts does not consider the heterogeneity of a HIN. In this paper, we use PCRW, which extends random walk based proximity to be applied for HINs. Second, as pointed out in \cite{line}, DeepWalk can only preserve second-order proximity, leading to poor performance in some tasks, such as link recover and classification, which require first-order proximity to be well-preserved. 


LINE \cite{line} is a recently proposed embedding approach for large-scale networks. Although it uses an explicit objective function to preserve the network structure, its performance suffers from the way it learns the vector representations. By design, LINE learns two representations separately, one preserving first-order proximity and the other preserving second-order proximity. Then, it directly concatenates the two vectors to form the final representation. 
In this paper, we propose a network embedding method, which does not distinguish the learning of first and second-order proximities and embeds proximities of all orders simultaneously. 

GraRep \cite{grarep} further extends DeepWalk to utilize high-order proximities.
GraRep does not scale well in large networks due to the expensive computation of the power of a matrix and the involvement of SVD in the learning process.
SDNE \cite{deep} is a semi-supervised deep model that captures the non-linear structural 
information over the network. The source code of SDNE is not available, so this approach cannot be reproduced and compared to ours.
Similarly, \cite{bordes2011learning} embeds entities in knowledge bases using an innovative neural network architecture and TriDNR \cite{tri} extends this embedding model to consider features from three aspects of the network: 1) network structure, 2) node content, and 3) label information. Our current work focuses on meta path based proximities, so it does not consider any other information in the HIN besides the network structure and the types of nodes and edges.   



\section{Problem Definition} \label{problemdefinition}

In this section, we formulate the problem of heterogeneous information network (HIN) embedding for meta path based proximities. We first define a HIN as follows:

\begin{definition}
\textbf{(Heterogeneous Information Network)} A heterogeneous information network (HIN) is a directed graph $G = (V, E)$ with an object type mapping function $\phi:V \rightarrow \mathcal{L}$ and a link type mapping function $ \psi:E \rightarrow \mathcal{R}$, where each object $v \in V$  belongs to an object type $\phi(v) \in \mathcal{L}$, and each link $e \in E$ belongs to a link type $\psi(e) \in \mathcal{R}$.  
\end{definition}

Figure \ref{fig:hinExp} illustrates a small  bibliographic HIN. We can see that the HIN contains two authors, three papers, three venues, two topics, and four types of links connecting them. 
For the ease of discussion, we assume that each object belongs to a single type; 
however, our technique can easily be adapted to the case where objects belong to multiple types.

\begin{definition}
\textbf{HIN Schema}. Given a HIN $G=(V,E)$ with mappings $ \phi:V \rightarrow \mathcal{L}$ and $ \psi:E \rightarrow \mathcal{R}$, the schema $T_G$ of $G$ is a directed graph defined over object types $\mathcal{L}$ and link types $\mathcal{R}$, i.e., $T_G=(\mathcal{L},\mathcal{R})$. 
\end{definition}

\begin{figure}
\subfigure[DBLP] { \label{fig:dblpSchema}
        \centering
        \includegraphics[width = 0.22\textwidth]{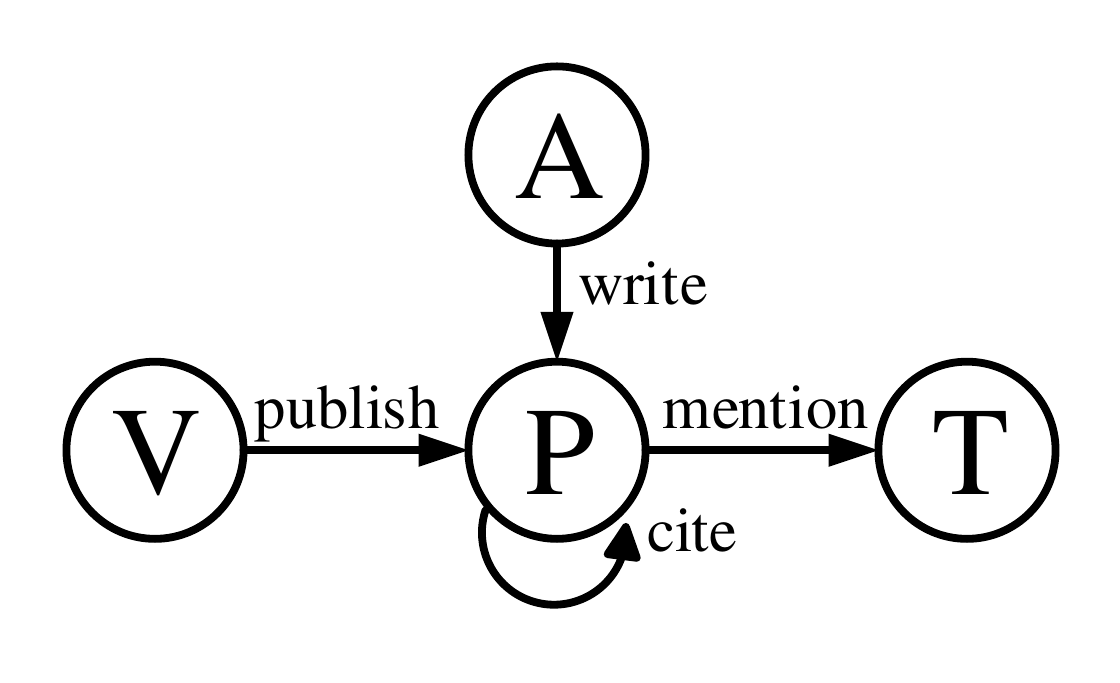}}	
\subfigure[MOVIE]{ \label{fig:movieSchema}
        \centering
       \includegraphics[width = 0.22\textwidth]{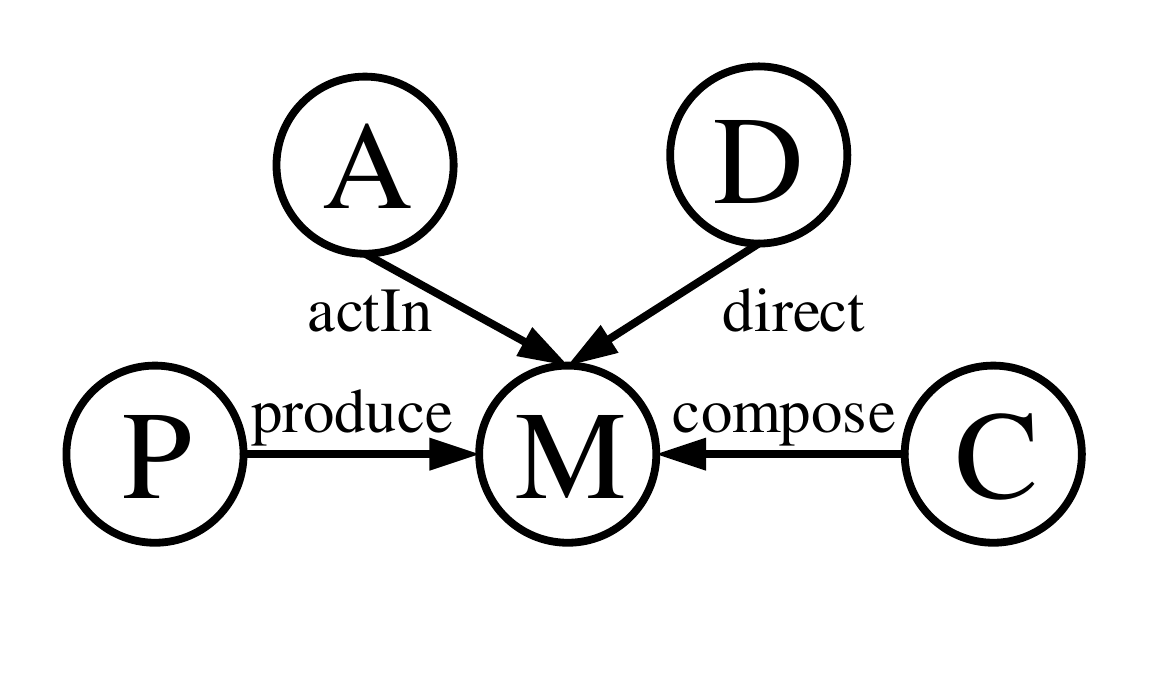}}	

\caption{Schemas of two heterogeneous networks}\label{fig:schema} 
\end{figure}

The HIN schema expresses all allowable link types between object types. Figure \ref{fig:dblpSchema} shows the schema of the HIN in Figure \ref{fig:hinExp}, where nodes $A$, $P$, $T$ and $V$ correspond to author, paper, topic and venue, respectively. There are also different edge types in the schema, such as `\emph{publish}' and `\emph{write}'. Figure \ref{fig:movieSchema} shows the schema of a movie-related HIN, with $M$, $A$, $D$, $P$, $C$ representing movie, actor, director, producer and composer, respectively.

In a HIN $G$, two  objects $o_1$, $o_2$ may be connected via multiple edges or paths.  Conceptually, each of these paths represents a specific direct or composite relationship between them. In Figure \ref{fig:hinExp}, authors $a_1$ and $a_2$ are connected via multiple paths. For example, $a_1\rightarrow p_3 \rightarrow a_2$ means that $a_1$ and $a_2$ are coauthors of paper $p_3$, and $a_1 \rightarrow p_1 \rightarrow p_2 \rightarrow a_2$ means that $a_2$'s paper $p_2$ cites $a_1$'s paper $p_1$. These two paths represent two different relationships between author $a_1$ and $a_2$.

Each different type of a relationship is modeled by a 
 \emph{meta path} \cite{pathsim}. 
Specifically, a meta path $\mathcal{P}$ is a sequence of object types $l_1, \dots, l_n$ connected by link types $r_1, \dots,  r_{n-1}$ as follows: 
$$\mathcal{P} = l_1 \xrightarrow[]{r_1} l_2  \dots l_{n-1} \xrightarrow[]{r_{n-1}} l_n.$$  For example, a meta path $A \xrightarrow[]{write} P \xrightarrow[]{write^{-1}} A$ represents the coauthor relationship between two authors.

An {\it instance} of the meta path $\mathcal{P}$ is a path in the HIN, which conforms to the pattern of $\mathcal{P}$. For example, path $a_1\rightarrow p_3 \rightarrow a_2$ in Figure \ref{fig:hinExp} is an instance of meta path $A \xrightarrow[]{write} P \xrightarrow[]{write^{-1}} A$. 

\begin{definition} \label{def:metapathproximity}
\textbf{(Meta Path based Proximity)} Given a HIN $G = (V, E)$ and a meta path $\mathcal{P}$, the proximity 
of two nodes $o_s,o_t\in V$ with respect to $\mathcal{P}$ is defined as:
\begin{equation}
s(o_s, o_t~|~\mathcal{P}) = \sum_{p_{o_s \rightarrow o_t} \in \mathcal{P}}{s(o_s, o_t~|~p_{o_s \rightarrow o_t} )}
\end{equation}
\noindent where $p_{o_s \rightarrow o_t}$ is a meta path instance of $\mathcal{P}$ linking from $o_s$ to $o_t$, and $s(o_s, o_t~|~p_{o_s \rightarrow o_t})$ is a proximity score w.r.t. the instance $p_{o_s \rightarrow o_t}$. 
\end{definition}

There are different ways to define $s(o_s, o_t~|~p_{o_s \rightarrow o_t})$ in the literature. Here, we only list two definitions that we use as test cases in our experiments.
 
\begin{itemize}
\item 
According to the \textbf{PathCount (PC)} model,
each meta path instance is equally important and should be given equal weight.
Hence, $s(o_s, o_t~|~p_{o_s \rightarrow o_t})=1$ and
the proximity between two objects w.r.t. $\mathcal{P}$ equals the number of instances of $\mathcal{P}$ connecting them, i.e., $s(o_s, o_t~|
~\mathcal{P}) = |\{p_{o_s \rightarrow o_t} \in \mathcal{P}\}|$.%
\footnote{For the ease of presentation, we overload notation $\mathcal{P}$ to also denote the set of instances of meta path $\mathcal{P}$.}

\item \textbf{Path Constrained Random Walk (PCRW)} \cite{pcrw} is a more sophisticated way to define the proximity $s(o_s, o_t~|~p_{o_s \rightarrow o_t})$ based on an instance $p_{o_s \rightarrow o_t}$. 
According to this definition, $s(o_s, o_t~|~p_{o_s \rightarrow o_t})$ is 
the probability that a random walk restricted on $\mathcal{P}$ would follow the instance $p_{o_s \rightarrow o_t}$.
\end{itemize}

Note that the embedding technique we introduce in this paper can also be easily adapted to other meta path based proximity measures, e.g., PathSim \cite{pathsim}, HeteSim \cite{hetesim} and BPCRW \cite{fspg}. 

\begin{definition} \label{def:proximity}
\textbf{(Proximity in HIN)} 
For each pair of objects $o_s, o_t \in V$, the proximity between  $o_s$ and $o_t$ in $G$ is defined as:
\begin{equation}
s(o_s, o_t) = \sum_{\mathcal{P}}{s(o_s, o_t~|~\mathcal{P})}
\end{equation}
\noindent where $\mathcal{P}$ is some meta path, and $s(o_s, o_t~|~\mathcal{P})$ is the proximity w.r.t. the meta path $\mathcal{P}$ as defined in Definition \ref{def:metapathproximity}. 
\end{definition}

According to Definition \ref{def:proximity}, the proximity of two objects equals the sum of the proximities w.r.t. all meta paths. Intuitively, this can capture all kinds of relationships between the two objects. We now provide a definition for the problem that we study in this paper.





\begin{definition}
\textbf{(HIN Embedding for Meta Path based Proximity)} 
Given a HIN $G = (V, E)$, 
develop a mapping $f: V \rightarrow R^d$ that transforms each object $o \in V$ to a vector in $R^d$, 
such that the proximities between any two objects in the original HIN are preserved in $R^d$.
\end{definition}

\section{\name} \label{method}

In this section, we introduce our methodology for embedding HINs. We
first discuss how to calculate the truncated estimation of meta
path based proximity in Section \ref{sec:proximitycalculation}. Then,
we introduce our model and define the objective function we want to
optimize in Section \ref{sec:model}. Finally, we present a negative
sampling approach that accelerates the optimization of the objective
function in Section \ref{sec:negative}.

\subsection{Truncated Proximity Calculation} \label{sec:proximitycalculation}

According to Definition \ref{def:proximity}, in order to compute the
proximity of two objects $o_i, o_j \in V$, we need to accumulate 
the corresponding meta path based proximity w.r.t. each meta path $\mathcal{P}$. 
For example, in Table \ref{table:instances}, we list some meta paths that have instances connecting $a_1$ and $a_2$ in Figure \ref{fig:hinExp}. We can see that there is one length-2 meta path, one length-3 meta path and two length-4 meta paths that have instances connecting $a_1$ and $a_2$. 
Generally speaking, the number of possible meta paths grows exponentially with their length and can be infinite for certain HIN schema (in this case, the computation of
 $s(o_i, o_j)$ is infeasible).

\begin{table}[] \small
\centering
\caption{Meta paths and instances connecting $a_1$ and $a_2$}
\label{table:instances}
\begin{tabular}{|c|c|c|c|c|}
\hline
Length             & $\mathcal{P}$& $p_{o_s \rightarrow o_t}$                                             & $s_{PC}$ & $s_{PCRW}$ \\ \hline
2                  & $APA$ &$a_1 \rightarrow p_3 \rightarrow a_2$                                            &  1       &  0.25      \\ \hline
3                  & $APPA$ & $a_1 \rightarrow p_1 \rightarrow p_2 \rightarrow a_2$                 &  1       &   0.5      \\ \hline
\multirow{3}{*}{4} & \multirow{2}{*}{$APTPA$} & $a_1 \rightarrow p_1 \rightarrow t_1 \rightarrow p_2\rightarrow a_2$  &  1       &0.25        \\ \cline{3-5} 
                   & & $a_1 \rightarrow p_3 \rightarrow t_2 \rightarrow p_3 \rightarrow a_2$ &  1       & 0.25       \\ \cline{2-5} 
                   & $APVPA$ & $a_1 \rightarrow p_3 \rightarrow v_3 \rightarrow p_3 \rightarrow a_2$ &   1      & 0.25       \\ \hline
$\cdots$           & $\cdots$ & $\cdots$                                                              & $\cdots$ &$\cdots$    \\ \hline
\end{tabular}
\end{table}

As pointed out in \cite{pathsim}, shorter meta paths are more informative than longer ones, because longer meta paths link more remote objects (which are less related semantically). 
Therefore, we use a  truncated estimation of proximity, which
only considers meta paths up to a length threshold $l$.
 This is also consistent with previous works on network embedding, which aim at reserving low-order proximity (e.g., \cite{deepwalk} reserves only second-order proximity, and \cite{line} first-order and second-order proximities).
Therefore, we define:
\begin{equation} \label{eq:proximity}
\hat{s}_{l}(o_i, o_j)=\sum_{len(\mathcal{P}) \le l}{s(o_i, o_j~|~\mathcal{P})}
\end{equation}
\noindent 
as the {\em truncated proximity}
between two objects $o_i$ and $o_j$.

For the case of PCRW based proximity, we have the following property:
\begin{property} \label{property}
$$\hat{s}_{l}(o_i, o_j)=\sum_{(o_i, o') \in E}{p_{o_i \rightarrow o'}^{\psi(o_i, o')} \times \hat{s}_{l-1}(o', o_j)}$$
\noindent where $p_{o_i \rightarrow o'}^{\psi(o_i, o')}$ is the transition probability from $o_i$ to $o'$ w.r.t. edge type $\psi(o_i, o')$. If there are $n$ edges from $o_i$ that belong to edge type $\psi(o_i, o')$, then $\psi(o_i, o') = \frac{1}{n}$.
\end{property}

\begin{proof}
Assume that $p = o_i \rightarrow o' \rightarrow \cdots \rightarrow o_j$ is a meta path instance of length $l$. According to definition of PCRW:
\begin{equation} \label{eq:proof}
s(o_i, o_j~|~p[1:l]) = p_{o_i \rightarrow o'}^{\psi(o_i, o')} \times s(o', o_j~|~p[2:l])
\end{equation}
\noindent where $p[i:j]$ is the subsequence of path instance $p$ from the $i$-th to the $j$-th objects, and $p[2:l]$ is a length $l-1$ path instance. Then, by summing over all the length-$l$ path instances for Equation \ref{eq:proof}, we get Property \ref{property}.
\end{proof}



\begin{algorithm} [] \small
	\caption{Calculate Truncated Proximity} \label{algorithm}
\KwIn { HIN $G = (V,E)$, length threshold $l$}
\KwOut{ Truncated Proximity Matrix $\hat{S}_l$}

	{$\hat{S}_{0} \leftarrow \emptyset$} \\
	\For {$o_s \in V$}{
		{$\hat{S}_{0}[o_s, o_s] \leftarrow 1.0$}\\
	}
	\For {$k \in [1 \cdots l]$}{
		{$\hat{S}_{k} \leftarrow \emptyset$}\\
		\For {$o_s \in V$}{
			\For {$o' \in neighbor(o_s)$}{
				\For {$(o', o_t) \in \hat{S}_{k-1}$}{
					{$\hat{S}_{k}[o_s, o_t] \leftarrow \hat{S}_{k}[o_s, o_t] + inc(o_s, o', \hat{S}_{k-1}[o', o_t])$} \\
				}
			}
		}
	}
	\Return {$\hat{S}_{l}$;} \\
\end{algorithm}

Based on Property \ref{property}, we develop a dynamic programming
approach (Algorithm \ref{algorithm}) to calculate the truncated
proximities. Basically, we compute the proximity matrix $\hat{S}_{k}$
for each $k$ from $1$ to $l$. We first initialize the proximity matrix
$\hat{S}_{0}$ (lines 1-3). Then, we use the transition function in
Property \ref{property} to update the proximity matrix for each $k$
(lines 4-9). If we use PCRW as the meta path based proximity measure, $inc(o_s, o', \hat{S}_{k-1}[o', o_t])$ in line 9 equals $p_{o_s \rightarrow o'}^{\psi(o_s, o')} \times \hat{S}_{k-1}[o', o_t]$. 
The algorithm can also be used for PathCount, in which case we set $inc(o_s, o', \hat{S}_{k-1}[o', o_t]) = \hat{S}_{k-1}[o', o_t]$.


We now provide a time complexity analysis for computing the truncated proximities on a HIN using Algorithm \ref{algorithm}. For each object $o_s \in V$, we need to enumerate all the meta path instances within length $l$. Suppose the average degree in the HIN is $D$; then, there are on average $l^D$ such instances. Hence, the total time complexity for proximity calculation is $O(|V| \cdot l^D)$, which is linear to the size of the HIN. 

\subsection{Model} \label{sec:model}

We now introduce our \name\ embedding method, which preserves the
meta path based proximities between objects as described above.
For each pair of objects $o_i$ and $o_j$, we use Sigmoid function to define their joint probability, i.e.,
\begin{equation}
p(o_i, o_j) = \frac{1}{1 + e^{- {v_i} \cdot {v_j}}}
\end{equation}
\noindent where ${v_i} (\text{or}~v_j) \in R^d$ is the low-dimensional representation of object $o_i$ (or $o_j$). $p: V^2 \rightarrow R$ is a probability distribution over a pair of objects in the original HIN.

In the original HIN $G$, the empirical joint probability of $o_i$ and $o_j$ can be defined as:
\begin{equation}
\hat{p}(o_i, o_j) = \frac{s(o_i, o_j)}{\sum_{o' \in V}{s(o_i, o')}}
\end{equation}

To preserve the meta path based proximity $s(\cdot,\cdot)$, a natural objective is to minimize the distance of these two probability distributions:
\begin{equation}
O = dist(\hat{p}, p)
\end{equation}

In this paper, we choose KL-divergence as the distance metric, so we have:
\begin{equation} \label{equation:objective}
O = -\sum_{o_i,o_j \in V}{s(o_i, o_j)\log{p(o_i, o_j)}}
\end{equation}





\subsection{Negative Sampling}\label{sec:negative}

Directly optimizing the objective function in Equation \ref{equation:objective} is problematic. First of all, there is a trivial solution: $v_{i,d} = \infty$ for all $i$ and all $d$. Second, it is computationally expensive to calculate the gradient, as we need to sum over all the non-zero proximity scores ${s}(o_i, o_j)$ for a specific object $o_i$. To address these problems, we adopt negative sampling proposed in \cite{mikolov2013distributed}, which basically samples a small number of negative objects to enhance the influence of positive objects.

Formally, we define an objective function for each pair of objects with non-zero meta path based proximity $s(o_i, o_j)$:

\begin{equation} \label{equation:sampling}
T(o_i, o_j) = -\log{(1 + e^{-v_i v_j})} - \sum_{1}^{K}{\mathbb{E}_{v' \in P_n(o_i)}[\log{(1+e^{v_i v'})}]}
\end{equation}
\noindent where $K$ is the times of sampling, and $P_n(v)$ is some noise distribution. As suggested by \cite{mikolov2013distributed}, we set $P_n(v) \propto d_{out}(v)^{3/4}$ where $d_{out}(v)$ is the out-degree of $v$.

We adopt the asynchronous stochastic gradient descent (ASGD) algorithm \cite{recht2011hogwild} to optimize the objective function in Equation \ref{equation:sampling}. Specifically, in each round of the optimization, we sample some pairs of objects $o_i$ and $o_j$ with non-zero proximity $s(o_i, o_j)$. Then, the gradient w.r.t. $v_i$ can be calculated as:

\begin{equation}
\frac{\partial{O}}{\partial{v_i}} = -s(o_i, o_j) \cdot \frac{e^{-v_i v_j}}{1 + e^{-v_iv_j}} \cdot v_j
\end{equation}
\section{Experiments} \label{section:experiments}
In this section, we conduct extensive experiments in order to test the effectiveness of our proposed HIN embedding approach. We first introduce our datasets and the set of methods to be compared in Section \ref{sec:configuration}. Then, we evaluate the effectiveness of all approaches on five important data mining tasks: network recovery (Section \ref{sec:networkrecover}), classification (Section \ref{sec:classification}), clustering (Section \ref{sec:clustering}), $k$-NN search (Section \ref{sec:knnsearch}) and visualization (Section \ref{sec:visulization}). 
In addition, we conduct a case study (Section \ref{sec:casestudy}) to compare the quality of top-$k$ lists. We evaluate the influence of parameter $l$ in Section \ref{sec:changel}.
Finally, we assess the runtime cost of applying our proposed transformation in Section \ref{sec:efficiency}.

\subsection{Dataset and Configurations} \label{sec:configuration}

\begin{figure}

\subfigure[YELP] { \label{fig:yelpSchema}
        \centering
        \includegraphics[width = 0.22\textwidth]{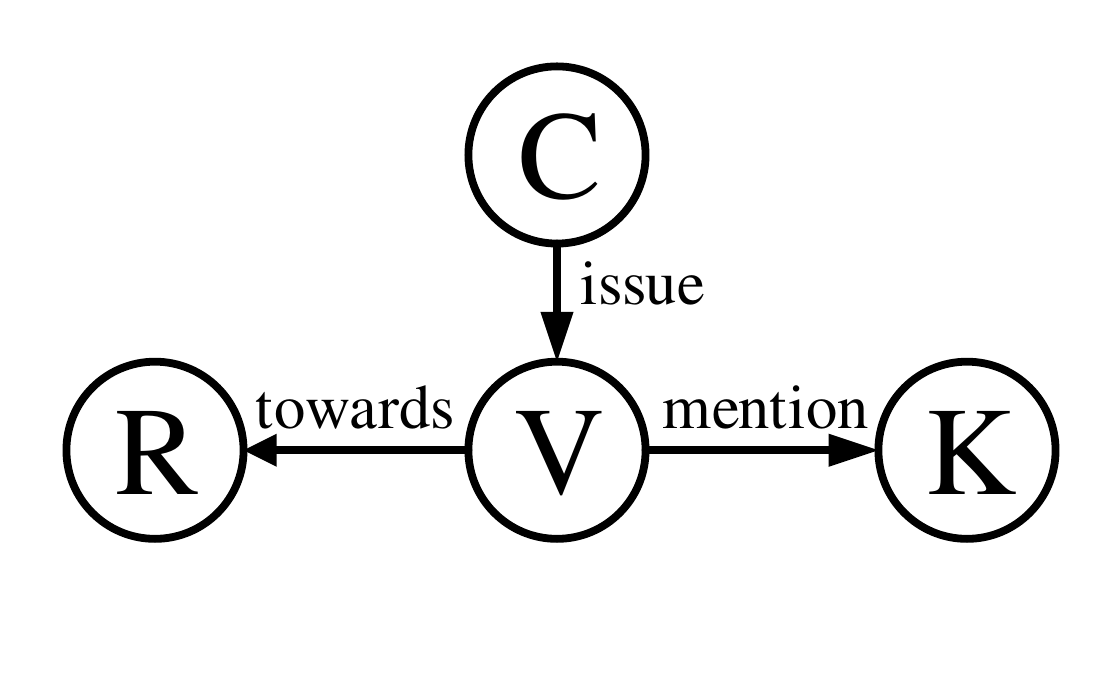}}	
\subfigure[GAME]{ \label{fig:gameSchema}
        \centering
       \includegraphics[width = 0.22\textwidth]{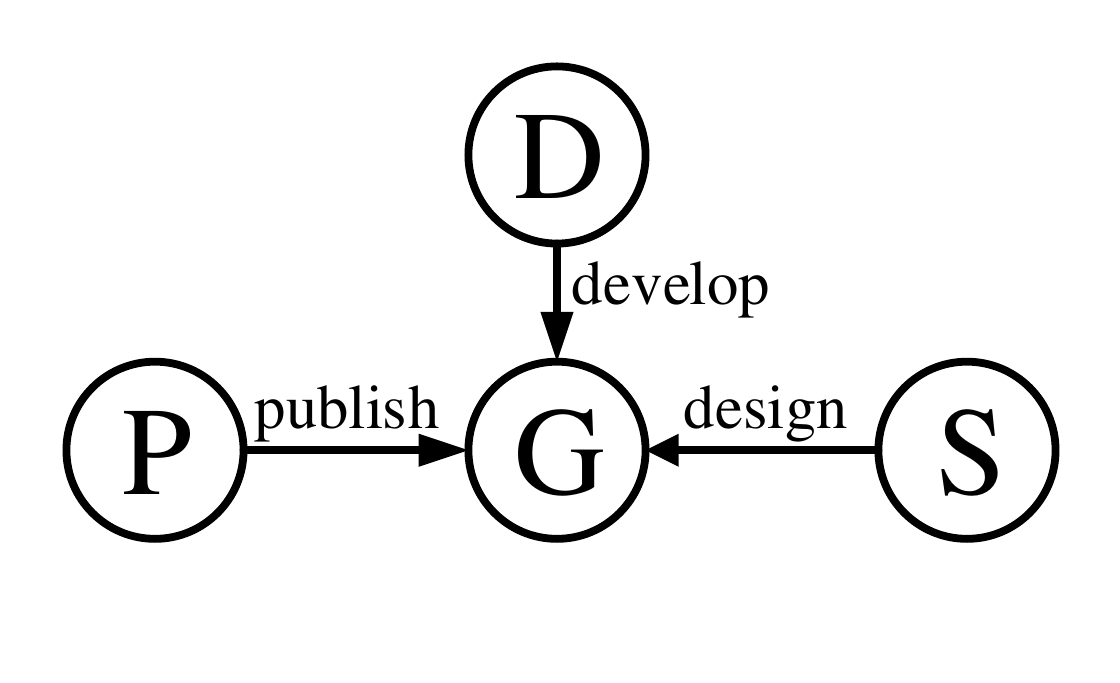}}	

\caption{Schemas of Two HINs}\label{fig:schema2} 
\end{figure}

\vspace{1em}
\noindent \textbf{Datasets}. We use four real datasets in our evaluation. Table \ref{table:statistics}  shows some statistics about them.

\begin{itemize}

\item \textbf{DBLP}. The schema of DBLP network is shown in Figure \ref{fig:dblpSchema}. We use a subset of DBLP, i.e., DBLP-4-Area taken of \cite{pathsim}, which contains 5,237 papers (P), 5,915 authors (A), 18 venues (V), 4,479 topics (T). The authors are from 4 areas: \emph{database}, \emph{data mining}, \emph{machine learning} and \emph{information retrieval}.

\item \textbf{MOVIE}. We extracted a subset from YAGO \cite{metastructure}, which contains knowledge about movies. The schema of MOVIE network is shown in Figure \ref{fig:movieSchema}. It contains 7,332 movies (M), 10,789 actors (A), 1,741 directors (D), 3,392 producers (P) and 1,483 composers (C). The movies are divided into five genres: \emph{action}, \emph{horror}, \emph{adventure}, \emph{sci-fi} and \emph{crime}. 

\item \textbf{YELP}. We extracted a HIN from YELP, which is about reviews given to  restaurants. It contains 33,360 reviews (V), 1,286 customers (C), 82 food-related keywords (K) and 2,614 restaurants (R). We only extract restaurants from one of the following types: \emph{fast food}, \emph{American}, \emph{sushi bar}. 
 The schema of the HIN is shown in Figure \ref{fig:yelpSchema}.

\item \textbf{GAME}. We extracted from Freebase \cite{bollacker2008freebase} a HIN, which is related to video games. The HIN consists of 4,095 games (G), 1,578 publishers (P), 2,043 developers (D) and 197 designers (S). All the game objects extracted are of one of the three genres: \emph{action}, \emph{adventure}, and \emph{strategy}. The schema is shown in Figure \ref{fig:gameSchema}.

\end{itemize}

\begin{table}[] 
\centering
\caption{Statistics of Datasets}
\label{table:statistics}
\begin{tabular}{|c|c|c|c|c|c|}
\hline
      & $|V|$  & $|E|$  & Avg. degree & $|\mathcal{L}|$ & $|\mathcal{R}|$ \\ \hline
DBLP  & 15,649 & 51,377 & 6.57        & 4              & 4            \\ \hline
MOVIE & 25,643 & 40,173 & 3.13        & 5              & 4            \\ \hline
YELP & 37,342 & 178,519 & 9.56        & 4              & 3            \\ \hline
GAME & 6,909  & 7,754   & 2.24        & 4              & 3            \\ \hline
\end{tabular}
\end{table}

\noindent \textbf{Competitors}. We compare the following network embedding approaches:

\begin{itemize}
\item \textbf{DeepWalk} \cite{deepwalk} is a recently proposed social network embedding method (see Section \ref{sec:related:embed} for details).
In our experiment settings, we ignore the heterogeneity and directly feed the HINs for embedding. We use default training settings, i.e., window size $w=5$ and length of random walk $t = 40$.

\item \textbf{LINE} \cite{line} is a method that preserves first-order and second-order proximities  between vertices (see Section \ref{sec:related:embed} for details). For each object, it computes two vectors; one for the first-order and one for the second-order proximities separately and then concatenates them. We use equal representation lengths for the first-order and second-order proximities, and use the default training settings; i.e., number of negative samplings. $K = 5$, starting value of the learning rate $\rho_0 = 0.025$, and total number of training samplings
$T = 100M$. Same as DeepWalk, we directly feed the HINs for embedding. 



\item \textbf{\name\_PC} is our \name\ model using PathCount as the meta path based proximity. We implemented the process of proximity computing in C++ on a 16GB memory machine with Intel(R) Core(TM) i5-3570 CPU @ 3.4 GHz. By default, we use $l = 2$. In Section \ref{sec:changel}, we study the influence of parameter $l$.

\item \textbf{\name\_PCRW} is our \name\ model using PCRW as the meta path based proximity. All the other configurations are the same as those of \name\_PC.

\end{itemize}

Unless otherwise stated, the dimensionality $d$ of the embedded vector space equals 10. 

\subsection{Network Recovery} \label{sec:networkrecover}

We first compare the effectiveness of different network embedding methods at a task of link recovery.
For each type of links (i.e., edges)
in the HIN schema, we enumerate all 
pairs of objects $(o_s, o_t)$ that can be connected by such a link 
and
calculate their proximity
in the low-dimensional space after embedding $o_s$ to $v_s$ and $o_t$ to $v_t$. 
Finally, we use the area under ROC-curve (AUC) to evaluate the performance of each embedding. 
For example, for edge type {\it write}, we enumerate all pairs of authors $a_i$ and papers $p_j$ in DBLP and compute the proximity for each pair.
Finally, using the real DBLP network as ground-truth, we compute the AUC value for each embedding method.

\begin{table*}[]
\centering
\caption{Accuracy of Network Recovery ($d = 10$)}
\label{table:networkrecover}
\begin{tabular}{c|c|c|c|c|c|c|c|}
\cline{2-8}
                                             & Edge Type & LINE\_1st & LINE\_2nd & LINE & DeepWalk & \name\_PC & \name\_PCRW \\ \hline
\multicolumn{1}{|c|}{\multirow{4}{*}{DBLP}}  & {\it write}     & 0.9885    & 0.7891    & 0.8907     & \textbf{0.9936}   & 0.9886    & 0.9839      \\ \cline{2-8} 
\multicolumn{1}{|c|}{}                       & {\it publish}   & 0.9365    & 0.6572    & 0.8057     & 0.9022   & 0.9330    & \textbf{0.9862}      \\ \cline{2-8} 
\multicolumn{1}{|c|}{}                       & {\it mention}   & \textbf{0.9341}    & 0.5455    & 0.7186     & 0.9202   & 0.8965    & 0.8879      \\ \cline{2-8} 
\multicolumn{1}{|c|}{}                       & {\it cite}      & \textbf{0.9860}    & 0.9071    & 0.9373     & 0.9859   & 0.9598    & 0.9517      \\ \hline
\multicolumn{1}{|c|}{\multirow{4}{*}{MOVIE}} & {\it actIn}     & 0.9648    & 0.5214    & 0.8425     & 0.7497   & 0.9403    & \textbf{0.9733}      \\ \cline{2-8} 
\multicolumn{1}{|c|}{}                       & {\it direct}    & 0.9420    & 0.5353    & 0.8468     & 0.7879   & 0.9359    & \textbf{0.9671}      \\ \cline{2-8} 
\multicolumn{1}{|c|}{}                       & {\it produce}    & 0.9685    & 0.5334    & 0.8599     & 0.7961   & 0.9430    & \textbf{0.9900}      \\ \cline{2-8} 
\multicolumn{1}{|c|}{}                       & {\it compose}   & 0.9719    & 0.5254    & 0.8574     & 0.9475   & 0.9528    & \textbf{0.9864}      \\ \hline
\multicolumn{1}{|c|}{\multirow{4}{*}{YELP}}  & {\it issue}     & 0.9249    & 0.4419    & 0.8744     & \textbf{1.0000}   & 0.5083    & 0.9960      \\ \cline{2-8} 
\multicolumn{1}{|c|}{}                       & {\it towards}    & 0.9716    & 0.4283    & 0.9221     & 0.5372   & 0.5155    & \textbf{0.9981}      \\ \cline{2-8} 
\multicolumn{1}{|c|}{}                       & {\it mention}    & \textbf{0.8720}    & 0.4178    & 0.7738     & 0.5094   & 0.5951    & 0.8534      \\ \hline
\multicolumn{1}{|c|}{\multirow{4}{*}{GAME}}  & {\it design}     & 0.9599    & 0.4997    & 0.6781     & {0.9313}   & 0.9219    & \textbf{0.9828}      \\ \cline{2-8} 
\multicolumn{1}{|c|}{}                       & {\it develop}    & 0.9021    & 0.5043    & 0.7736     & 0.9328   & 0.8592    & \textbf{0.9843}      \\ \cline{2-8} 
\multicolumn{1}{|c|}{}                       & {\it publish}    & {0.7800}    & 0.4719    & 0.6786     & 0.9711   & 0.7617    & \textbf{0.9756}      \\ \hline
\multicolumn{1}{|c|}{Overall}                & Average   & 0.9359    & 0.5556    & 0.8185     & 0.8546   & 0.8365    & \textbf{0.9655}      \\ \hline
\end{tabular}
\end{table*}

\begin{table*}[!h] \small
\centering
\caption{Top-5 similar lists for {\it Jiawei Han}}
\label{table:jiawei}
\begin{tabular}{|c|c|c|c|c|c|c|c|c|c|}
\hline
    & \multicolumn{3}{c|}{LINE}         & \multicolumn{3}{c|}{DeepWalk}       & \multicolumn{3}{c|}{\name\_PCRW}        \\ \hline
$k$ & conf  & author      & topic       & conf  & author         & topic      & conf  & author             & topic      \\ \hline
1   & PAKDD & Jiawei Han  & clustermap  & KDD   & Jiawei Han     & itemsets   & KDD   & Jiawei Han         & closed     \\ \hline
2   & PKDD  & Gao Cong    & orders      & PAKDD & Xifeng Yan     & farmer     & ICDM  & Xifeng Yan         & mining     \\ \hline
3   & SDM   & Lei Liu     & summarizing & PKDD  & Ke Wang        & closed     & SDM   & Ke Wang            & itemsets   \\ \hline
4   & KDD   & Yiling Yang & ccmine      & SDM   & Yabo Xu        & tsp        & PAKDD & Christos Faloutsos & sequential \\ \hline
5   & ICDM  & Daesu Lee   & concise     & ICDM  & Jason Flannick & prefixspan & PKDD  & Xiaoxin Yin        & massive    \\ \hline
\end{tabular}
\end{table*}

\begin{figure*}[!h]
     \subfigure[Avg. \#paper w.r.t. $k$]{ \label{fig:WWW}
         \centering
         \includegraphics[width=0.33\textwidth]{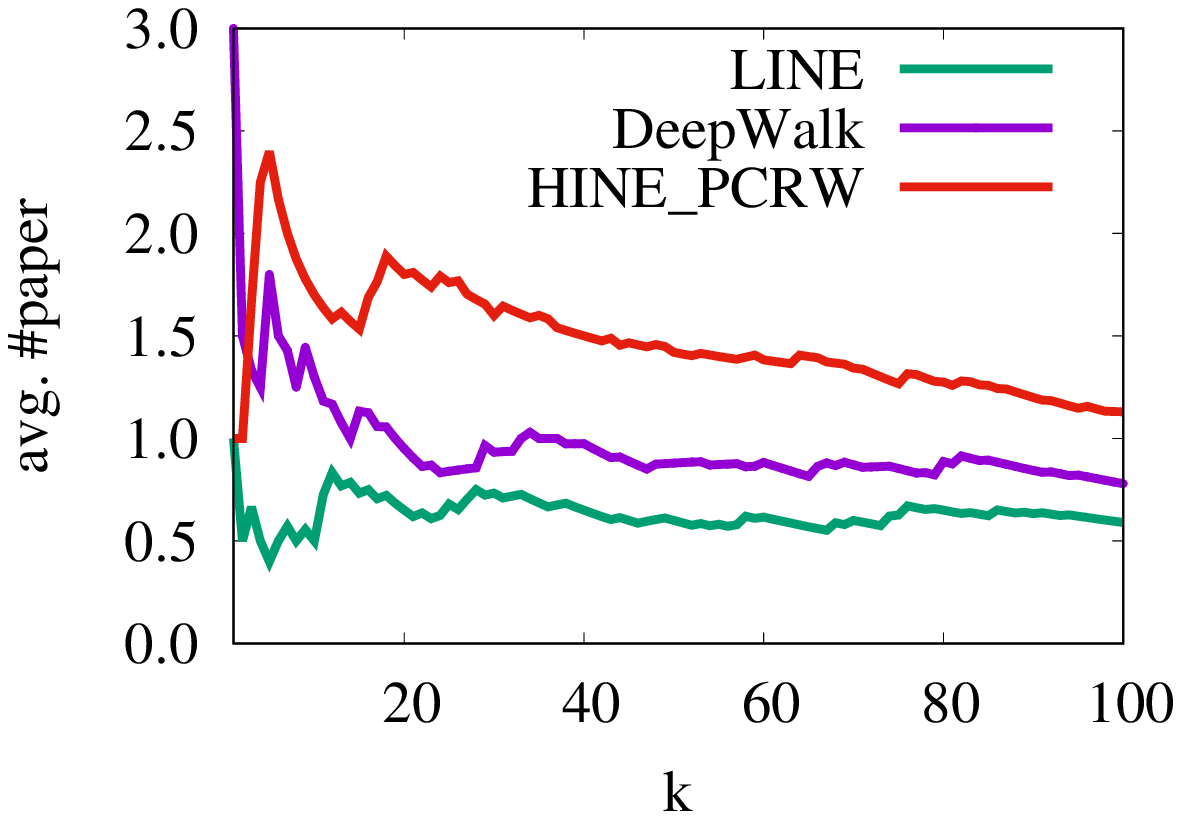}}
     \subfigure[$\mathcal{F}$ w.r.t. $k$]{ \label{fig:topklistsF} 
         \centering
         \includegraphics[width=0.33\textwidth]{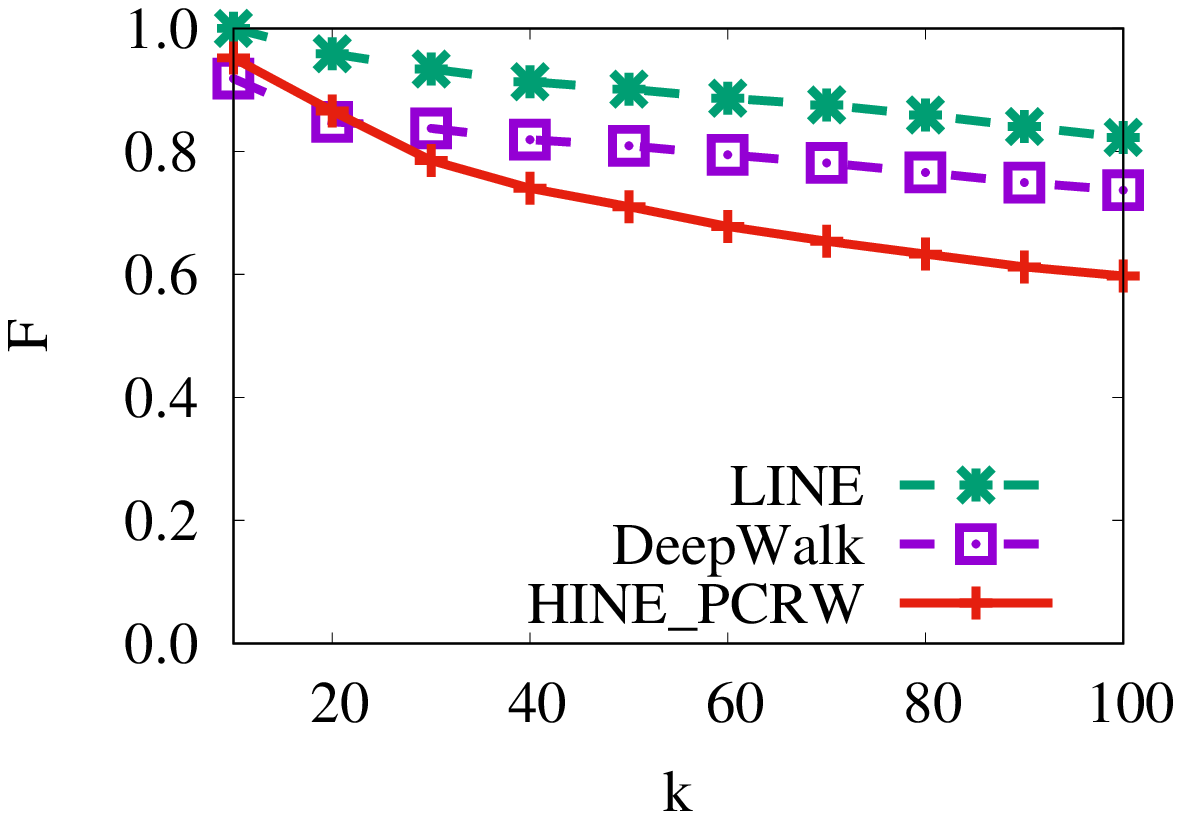}} 
     \subfigure[$\mathcal{K}$ w.r.t. $k$]{\label{fig:topklistsK} 
         \centering
         \includegraphics[width=0.33\textwidth]{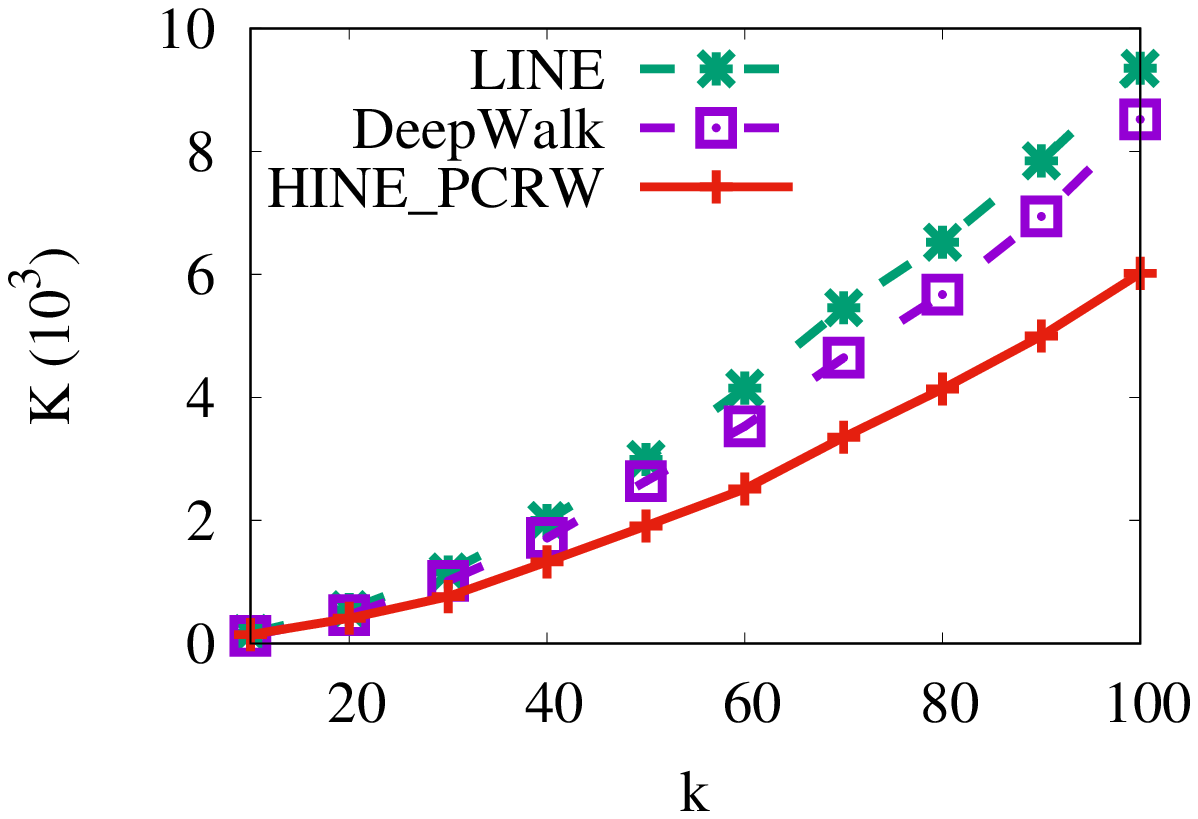}}
 \caption{Results of Top-$k$ lists for venue {\it WWW}}\label{fig:topklists} 
 \end{figure*}

The results for $d = 10$ are shown in Table \ref{table:networkrecover}. 
Observe that, in general, \name\_PC and \name\_PCRW have better  performance compared 
to LINE and DeepWalk. In order to analyze the reasons behind the bad performance of LINE, we also included two special versions of LINE: \textbf{LINE\_1st} (\textbf{LINE\_2nd}) is a simple optimization approach that just uses stochastic gradient descent to optimize just the first-order (second-order) proximity among the vertices.
LINE\_1st has much better performance than LINE\_2nd because second-order proximity preservation does not facilitate link prediction.
LINE, which concatenates LINE\_1st and LINE\_2nd vectors, has worse performance than LINE\_1st because its LINE\_2nd vector component harms link prediction accuracy.
This is consistent with our expectation, that training proximities of multiple orders simultaneously is better than training them separately. 
Among all methods, \name\_PCRW has the best performance in preserving the links of HINs; in the cases where   \name\_PCRW loses by other methods, its AUC is very close to them.
\name\_PCRW outperforms \name\_PC, which is consistent with results in previous work that find PCRW superior to PC (e.g., \cite{metastructure}). 
The rationale is that a random walk models proximity as probability,
which naturally weighs nearby objects higher compared to remote ones. 

\subsection{Classification} \label{sec:classification}

\begin{table}[!h] \small
\centering
\caption{Results of Classification with $d = 10$}
\label{table:classification}
\begin{tabular}{cc|c|c|c|c|}
\cline{3-6}
                                             &                & LINE  & DeepWalk & \name\_PC    &  \name\_PCRW           \\   \hline \hline
\multicolumn{1}{|l|}{\multirow{2}{*}{DBLP}}  &  Macro-F1      & 0.816 & 0.839    & 0.844 & \textbf{0.868} \\ \cline{2-6} 
\multicolumn{1}{|l|}{}                       & Micro-F1  & 0.817 & 0.840    & 0.844 & \textbf{0.869} \\ \hline \hline
\multicolumn{1}{|l|}{\multirow{2}{*}{MOVIE}} & Macro-F1     & 0.324 & 0.280    & 0.346 & \textbf{0.365} \\ \cline{2-6} 
\multicolumn{1}{|l|}{}                       & Micro-F1     & 0.369 & 0.328    & 0.387 & \textbf{0.406} \\ \hline \hline
\multicolumn{1}{|l|}{\multirow{2}{*}{YELP}} & Macro-F1       & \textbf{0.898} & 0.339    & 0.470 & {0.886} \\ \cline{2-6} 
\multicolumn{1}{|l|}{}                       & Micro-F1      & \textbf{0.887} & 0.433    & 0.518 & {0.878} \\ \hline \hline
\multicolumn{1}{|l|}{\multirow{2}{*}{GAME}} & Macro-F1       & {0.451} & \textbf{0.487}    & 0.415 & 0.438 \\ \cline{2-6} 
\multicolumn{1}{|l|}{}                       & Micro-F1       & {0.518} & \textbf{0.545}    & 0.501 & {0.518} \\ \hline \hline
\multicolumn{1}{|l|}{\multirow{2}{*}{\textbf{Overall}}} & Macro-F1    & 0.622 & 0.486    & 0.519 & \textbf{0.639} \\ \cline{2-6} 
\multicolumn{1}{|l|}{}                       & Micro-F1      & 0.648 & 0.537    & 0.563 & \textbf{0.668} \\ \hline
\end{tabular}
\end{table}

\begin{table}[!h]
\centering
\caption{Results of Clustering with $d = 10$}
\label{table:clustering}
\begin{tabular}{|c|c|c|c|c|}
\hline
             NMI               & LINE   & DeepWalk        & \name\_PC     & \name\_PCRW            \\ \hline \hline 
{DBLP}      & 0.3920 & \textbf{0.4896} & 0.4615 & 0.4870          \\ \hline
{MOVIE}    & 0.0233 & 0.0012          & \textbf{0.0460} & \textbf{0.0460} \\ \hline
{YELP}     & \textbf{0.0395} & 0.0004          & 0.0015 & {0.0121} \\ \hline
{GAME}    & {0.0004} & 0.0002          & \textbf{0.0022} & {0.0004} \\ \hline \hline
{\textbf{Overall}}     & 0.1138 & 0.1229          & 0.1278 & \textbf{0.1364} \\ \hline
\end{tabular}
\end{table}

       

 \begin{figure*}[!h]
  \includegraphics[width=\linewidth]{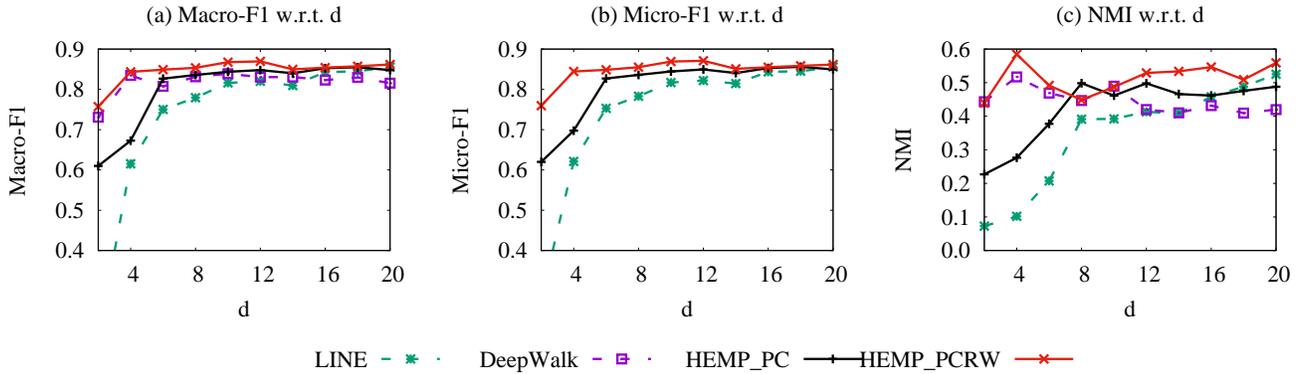}
  \caption{Performance of Classification and Clustering w.r.t. $d$}
  \label{fig:varyd}
\end{figure*}

\begin{figure}[!h]
		 \subfigure[LINE]{ 
         \centering
         \includegraphics[width=0.23\textwidth]{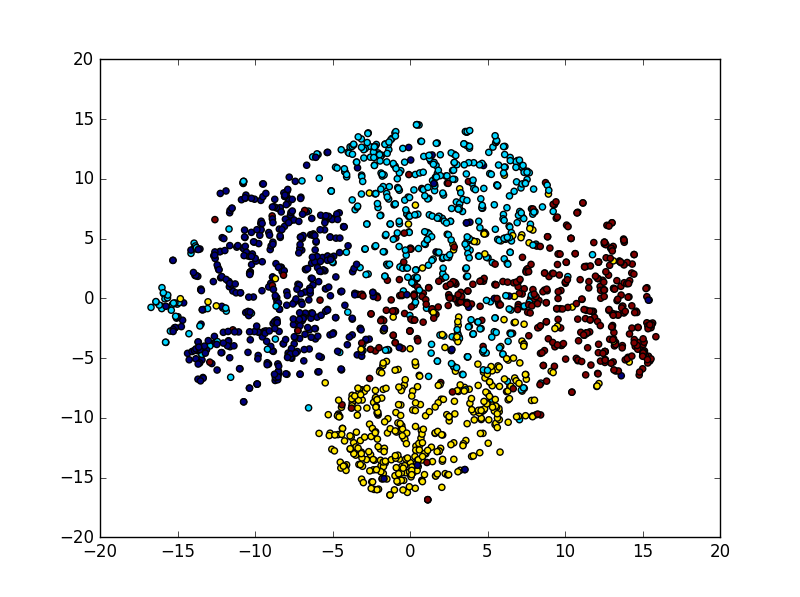}}	
		 \subfigure[DeepWalk]{ 
         \centering
         \includegraphics[width=0.23\textwidth]{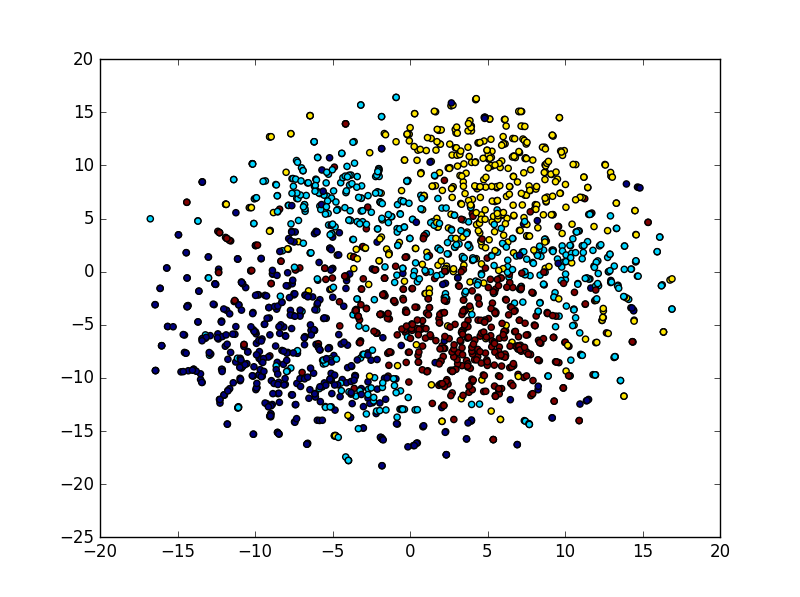}}	
		 \subfigure[\name\_PC]{
         \centering
         \includegraphics[width=0.23\textwidth]{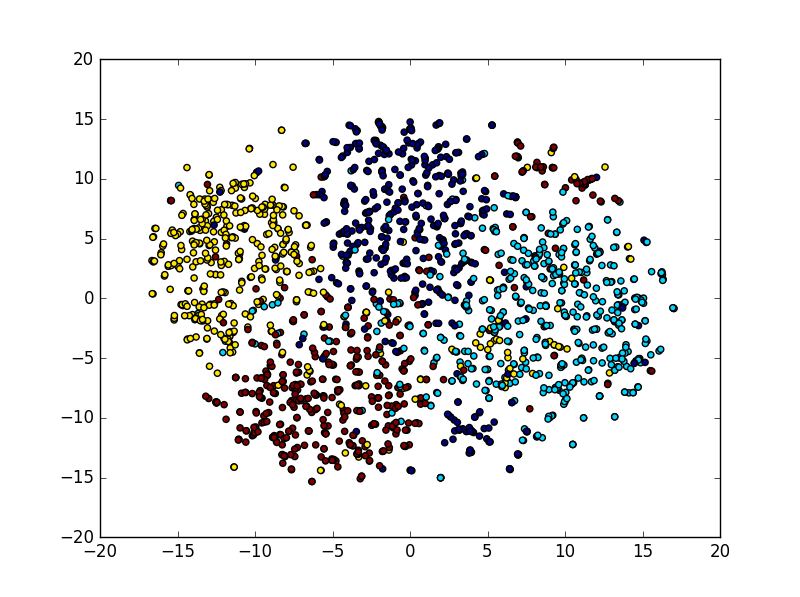}}
		 \subfigure[\name\_PCRW]{ 
         \centering
         \includegraphics[width=0.23\textwidth]{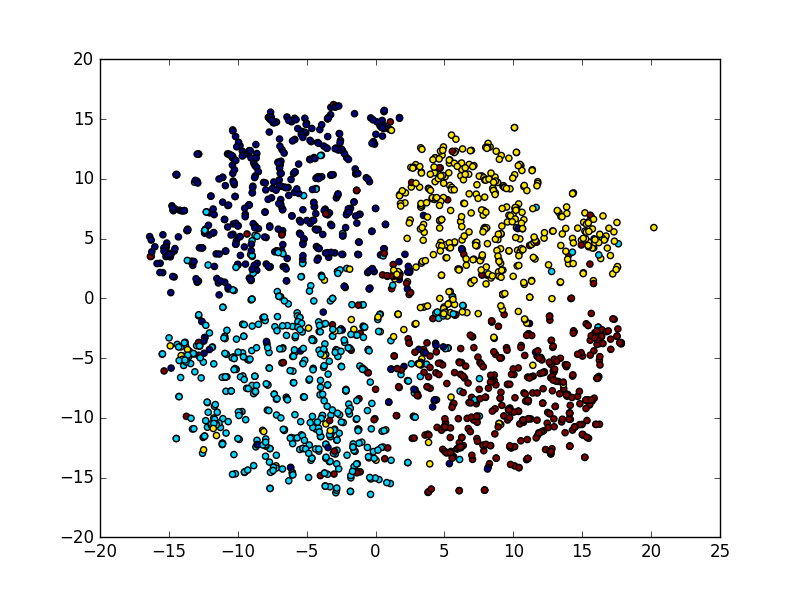}}	
 \caption{Visualization Results on DBLP}\label{fig:visualization} 
 \end{figure}

We conduct a task of multi-label classification. For DBLP, we use the areas of authors as labels. For MOVIE, we use the genres of movies as labels. For YELP, we use the restaurant types  as labels. For GAME, we use the type of games as labels. We first use different methods to embed the HINs.
Then, we randomly partition the samples into a training and a testing set with ratio $4:1$. Finally, we use $k$ nearest neighbor ($k-$NN) classifiers with $k=5$ to predict the labels of the test samples. We repeat the process for $10$ times and compute the average Macro-F1 and Micro-F1 scores to evaluate the performance. 

Table \ref{table:classification} shows the results for $d = 10$. We can see that all methods have better results on DBLP and YELP, than on MOVIE and GAME. This is because the average degree on MOVIE and GAME is smaller, and the HINs are sparser (See Table \ref{table:statistics}). Observe that \name \_PC and \name\_PCRW outperform DeepWalk and LINE in DBLP and MOVIE; \name\_PCRW performs the best in both datasets.
On YELP, LINE has slightly better results than \name\_PCRW (about
1\%), while DeepWalk has a very poor performance. On GAME, DeepWalk is
the winner, while LINE and \name\_PCRW perform similarly. Overall, we can see that \name\_PCRW has the best (or close to the best) performance, and LINE has quite good performance. On the other hand, DeepWalk's performance is not stable.

We also measured the performance of the methods for different values of $d$. Figures \ref{fig:varyd}(a) and \ref{fig:varyd}(b) show the results on DBLP. 
As $d$ becomes larger, all approaches get better in the beginning, and then converge to a certain performance level. Observe that overall, \name\_PCRW has the best performance in classification tasks.

\subsection{Clustering} \label{sec:clustering}

We also conducted clustering tasks to assess the performance of the methods. In DBLP we cluster the authors, in MOVIE we cluster the movies, in YELP we cluster the restaurants, and in GAME we cluster the games. We use the same ground-truth as in Section \ref{sec:classification}. We use normalized mutual information (NMI) to evaluate performance.

Table \ref{table:clustering} shows the results for $d=10$. 
We can see that the performance of all methods in clustering tasks is
not as stable as that in classification. All the embedding methods
perform well on DBLP, but they have a relatively bad performance on
the other three datasets. On DBLP, DeepWalk and \name\ have better
performance than LINE. On MOVIE, \name\_PC and \name\_PCRW outperform
all other approaches.  On YELP, LINE has 
better performance than \name\_PCRW, and DeepWalk has very poor
performance. On GAME, DeepWalk outperforms the others. 
Overall, we can see that \name\_PCRW outperforms all the other methods
in the task of clustering. 


Figure \ref{fig:varyd}(c) shows performance of the approaches for different $d$ values on DBLP. 
Observe that \name\_PCRW in general outperforms all other methods. 
Only for a narrow range of $d$ ($d=10$) DeepWalk slightly outperforms \name\_PCRW
(as also shown in Table \ref{table:clustering}).
Generally speaking, \name\_PCRW best preserves the proximities among authors in the task of clustering.



\subsection{k-NN Search} \label{sec:knnsearch}

We compare the performances of three methods, i.e., LINE, DeepWalk and \name\_PCRW, on $k$-NN search tasks. Specifically, we conduct a case study on DBLP, to compare the quality of $k$ nearest {\it authors} for venue {\it WWW} in the embedded space.
We first evaluate the quality of $k$-NN search by counting the average number of papers that the authors in the $k$-NN  result have published in WWW, when varying $k$ from 1 to 100 (Figure \ref{fig:WWW}).
We can see that the nearest authors found in the embedding by \name\_PCRW have more papers published in WWW compared to the ones found in the spaces of LINE and DeepWalk.

We then use the top-$k$ author list for venue WWW in the original DBLP network as ground-truth.
 We use two different metrics to evaluate the quality of top-$k$ lists in the embedded space, i.e., Spearman's footrule $\mathcal{F}$ and Kendall's tau $\mathcal{K}$ \cite{fagin2003comparing}. 
The results are shown in Figures \ref{fig:topklistsF} and \ref{fig:topklistsK}. We can see that the top-$k$ list of \name\_PCRW is closer to the one in the original HIN.

\subsection{Visualization} \label{sec:visulization}

We compare the performances of all approaches on the task of visualization, which aims to layout an HIN on a 2-dimensional space. Specifically, we first use an embedding method to map DBLP into a vector space, then, we map the vectors of authors to a 2-D space using the t-SNE \cite{maaten2008visualizing} package. 

The results are shown in Figure \ref{fig:visualization}.
LINE can basically separate the authors from different groups (represented by the same color), but some blue points mixed with other groups.
The result of DeepWalk is not very good, as many authors from different groups are mixed together. 
\name\_PC clearly has better performance than DeepWalk. Compared with LINE, \name\_PC can better separate different groups. Finally, \name\_PCRW's result is the best among these methods, because it clearly separates the authors from different groups, leaving a lot of empty space between the clusters.
This is consistent with the fact that \name\_PCRW has the best performance in classifying the authors of DBLP.

\subsection{Case Study} \label{sec:casestudy}

We perform a case study, which shows the $k$-NN objects to a given object in DBLP. Specifically, we show in Table \ref{table:jiawei} the top-5 closest {\it venues}, {\it authors} and {\it topics} to author {\it Jiawei Han} in DBLP. 

By looking at the results for top-$5$ venues, we can see that LINE does not give a good top-5 list, as it cannot rank KDD in the top publishing venues of the author. DeepWalk is slightly better, but it ranks ICDM at the 5th position, while \name\_PCRW ranks KDD and ICDM as 1st and 2nd, respectively. Looking at the results for authors, observe that \name\_PCRW gives a similar top-5 list to DeepWalk, except that \name\_PCRW prefers top researchers, e.g., Christos Faloutsos. Looking the results for topics, note that only \name\_PCRW can provide a general topic like ``mining''.

\subsection{Effect of the $l$ Threshold} \label{sec:changel}

\begin{figure}[]
		 \subfigure[Macro-F1 w.r.t. $l$]{ 
         \centering
         \includegraphics[width=0.23\textwidth]{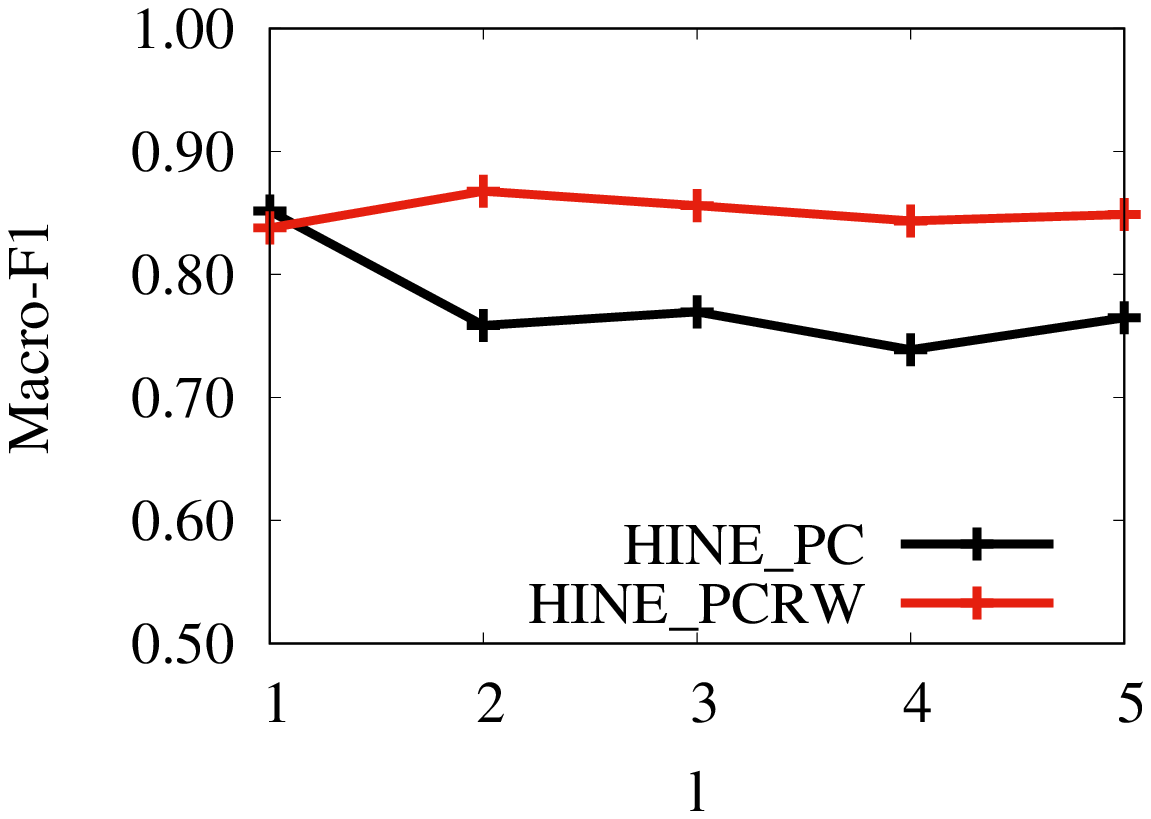}}	
		 \subfigure[Micro-F1 w.r.t. $l$]{
        \centering
         \includegraphics[width=0.23\textwidth]{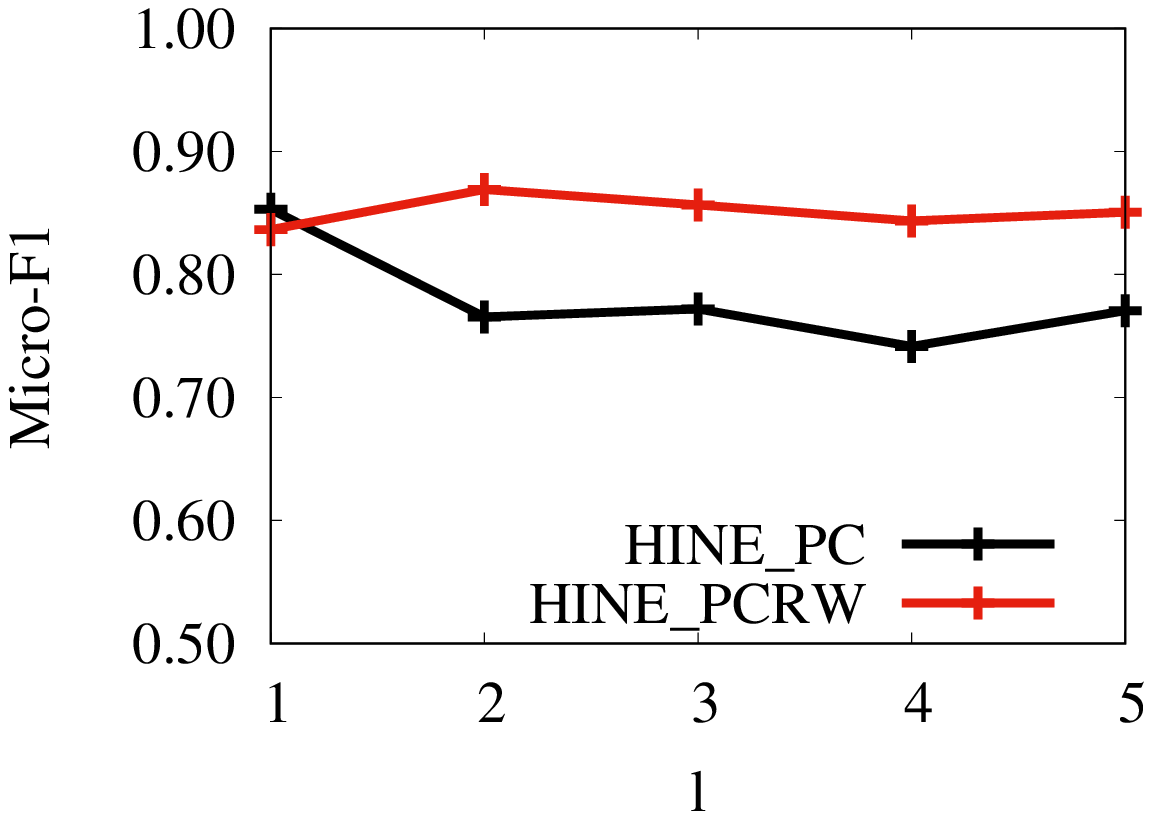}}	
		 \subfigure[NMI w.r.t. $l$]{ 
         \centering
         \includegraphics[width=0.23\textwidth]{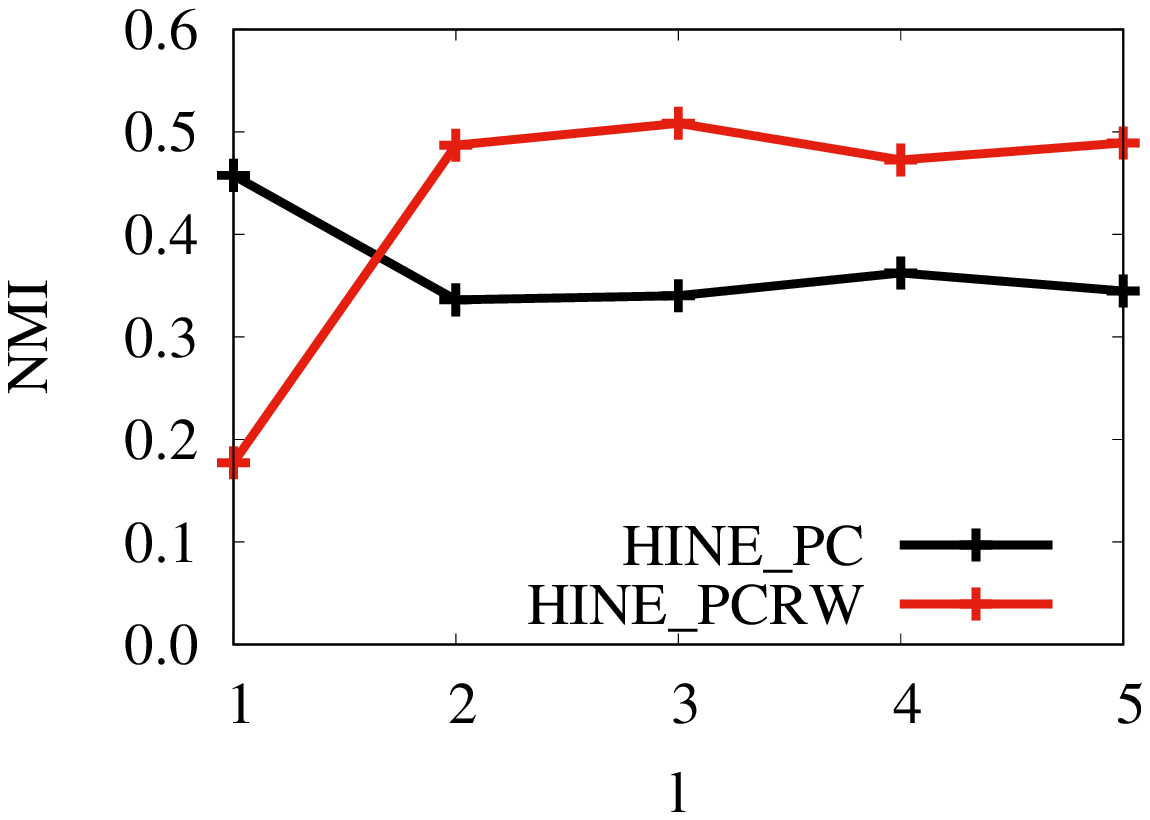}}		
        \subfigure[Execution time w.r.t. $l$]{ \label{fig:efficiency}
         \centering
         \includegraphics[width=0.23\textwidth]{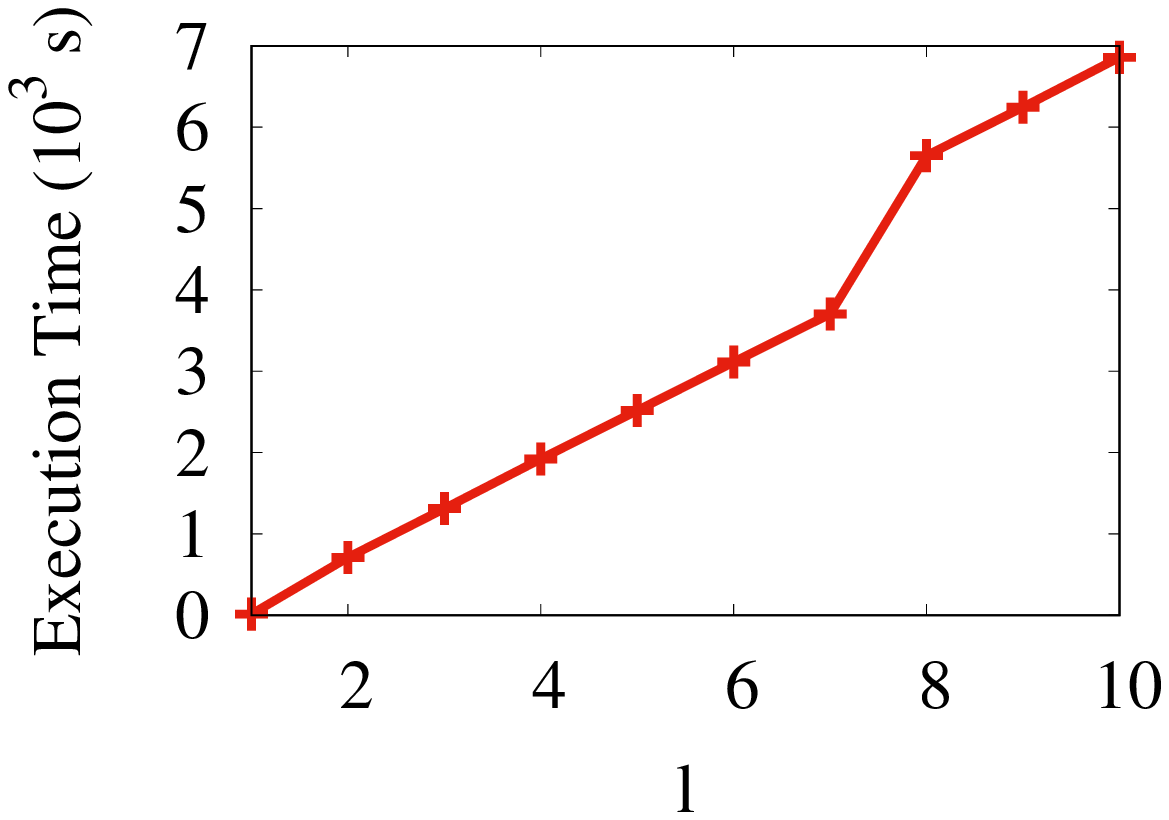}}
 \caption{Performances w.r.t. $l$}\label{fig:varyl} 
 \end{figure}

We now study the effect of $l$ on different data mining tasks, e.g., classification and clustering. 
 Figure \ref{fig:varyl}(a) and (b) shows the results of classification on DBLP. We can see that 1) \name\_PCRW has quite stable performance w.r.t. $l$, and it achieves its best performance when $l = 2$. 2) The performance of \name\_PC is best when $l = 1$. This is because PathCount views meta path instances of different length equally important, while PCRW assigns smaller weights to the longer instances when performing the random walk. This explains why \name\_PCRW outperforms \name\_PC in these tasks.
 
 Figure \ref{fig:varyl}(c) shows the results of clustering on DBLP. The results are similar those of classification, except that \name\_PC has much better performance than \name\_PCRW when $l = 1$. This is natural, because when $l=1$, \name\_PCRW  can only capture very local proximities. When $l>1$, \name\_PCRW outperforms \name\_PC.

 

\subsection{Efficiency} \label{sec:efficiency}



We show in Figure \ref{fig:efficiency} the running time for computing all-pair truncated proximities on DBLP with the method described in Section \ref{sec:proximitycalculation}. We can see that our proposed algorithm can run in a reasonable time and scales well with $l$. Specifically, in our experiments with $l=2$, the time for computing all-pair proximities is less than 1000s. 
In addition, note that using ASGD to solve our objective function in our experiments is very efficient, taking on average 27.48s on DBLP with $d = 10$.

\section{Conclusion} \label{conclusion}

In this paper, we study the problem of heterogeneous network embedding for meta path based proximity, which can fully utilize the heterogeneity of the network.
 We also define an objective function, which aims at minimizing the distance of two distributions, one modeling the meta path based proximities, the other modeling the proximities in the embedded low-dimensional space. We also investigate using negative sampling to accelerate the optimization process. As shown in our extensive experiments, our embedding methods can better recover the original network, and have better performances over several data mining tasks, e.g., classification, clustering and visualization, over the state-of-the-art network embedding methods. 



{  
\bibliographystyle{abbrv} \small
\bibliography{ref/refs}
}

\end{document}